# Few-shot Learning using Data Augmentation and Time-Frequency Transformation for Time Series Classification

Hao Zhang
School of Control Science and Engineering
Shandong University
Jinan, China
202000172001@mail.sdu.edu.cn

Zhendong Pang
School of Control Science and Engineering
Shandong University
Jinan, China
201800171089@mail.sdu.edu.cn

Jiangpeng Wang
School of Control Science and Engineering
Shandong University
Jinan, China
202000171026@mail.sdu.edu.cn

Teng Li*
School of Control Science and Engineering
Shandong University
Jinan, China
* Corresponding author: li.teng@sdu.edu.cn

*Abstract*—Deep neural networks (DNNs) that tackle the time series classification (TSC) task have provided a promising framework in signal processing. In real-world applications, as a data-driven model, DNNs are suffered from insufficient data. Few-shot learning has been studied to deal with this limitation. In this paper, we propose a novel few-shot learning framework through data augmentation, which involves transformation through the time-frequency domain and the generation of synthetic images through random erasing. Additionally, we develop a sequence-spectrogram neural network (SSNN). This neural network model composes of two sub-networks: one utilizing 1D residual blocks to extract features from the input sequence while the other one employing 2D residual blocks to extract features from the spectrogram representation. In the experiments, comparison studies of different existing DNN models with/without data augmentation are conducted on an amyotrophic lateral sclerosis (ALS) dataset and a wind turbine fault (WTF) dataset. The experimental results manifest that our proposed method achieves 93.75% F1 score and 93.33% accuracy on the ALS datasets while 95.48% F1 score and 95.59% accuracy on the WTF datasets. Our methodology demonstrates its applicability of addressing the few-shot problems for time series classification.

*Keywords-deep neural networks; few-shot learning; data augmentation; time series classification*

## I. INTRODUCTION

Time series signal (TSS) has been collected and analyzed in many real-world monitoring applications, such as precise medical disease diagnosis and the intricate task of anomaly detection within machinery and industrial systems. The two main attributes of TSS are sequential ordering and dynamics [1, 2], each of which has been delicately analyzed and studied by investigators through various methodologies.

Given the large amount of data involved in real-world data preprocessing, deep learning methods have been actively studied for TSC problems [3]. These methods aim to solve more difficult problems that are challenging to mine hidden, complex, and more discriminatory features within the data, leading to higher model performance. With the powerful capabilities in representation learning, end-to-end learning, and high performance, deep learning algorithms have achieved impressive successes in the field of TSC including residual network (ResNet), gate recurrent unit (GRU) and fully convolution networks (FCNs) [4-6]. However, to prepare a well-trained deep neural network, a sufficient dataset is in demand to prevent models from overfitting, a condition that can significantly impair their performance when faced with new test samples. As a matter of fact, datasets containing supervised information could be relatively elusive and exclusive due to some scenarios constrained by privacy, safety, ethical issues, or the intricacy of sampling operations.

To address this issue, researchers have focused on the problem of few-shot learning, which has gained increasing attention for handling the tasks involving datasets with scarce samples. Inspiringly, various data augmentation measures have emerged to mitigate overfitting problems, including approaches such as noise injection, simple transformations, meta-learning, generative adversarial networks [7, 8], and others. Distinguished from image data, augmentation of time-series data requires reasonable design since different types of TSSs contain different intrinsic information.

Straightforward methods of image augmentation measures have been widely adopted in the CNN images classification. For instance, the cropping and scaling in the original proposal of AlexNet or VGG [9, 10] can be implemented directly to enlarge the amount of image datasets. Three different TF approaches, the spectrogram of short-time Fourier transform (STFT), continuous wavelet transform and smoothed pseudo Wigner-Ville distribution have been adopted in [11], transforming the modality of time-series into images. However, normal augmentation of image transformation like cropping and scaling can barely be applied to time-series problems. This is due to the shape or the amplitude changes of these operations can drastically alter the ordered information and compromise the dynamics features of peculiar TSSs.

The contributions of this paper are highlighted as follows: (1) We implement data augmentation by TF transformation using STFT and image generation using random erasing, which addressing the few-shot learning problem of insufficient time series datasets. (2) We propose a novel DNN models called Sequence-Spectrogram Neural Network (SSNN), which can extract the latent features within both time and frequency domains. (3) We conduct the proposed model for both univariate and multivariate time series and validated the proposed model on real-world datasets to compare and validate the performance of the proposed model.

## II. PROBLEM DEFINE

The general goal of this work is to address the few-shot learning problem for TSC through deep learning. Specifically, we propose converting the TSS from whole series $X_l \in X$ to time-frequency spectrogram $Y_k \in Y$ through STFT function $S$, which serves to augment the time-series with more modalities:

$$Y_k = S(X_l) \quad (1)$$

Then the generated synthetic images with random erasing $R_j$ are zero in increasing the total amount of the TSS dataset $\hat{Y}$, where $j$ is defined as the yielding number of a single sample:

$$\hat{Y}_{k1}, \hat{Y}_{k2}, \ldots, \hat{Y}_{kj} = R(Y_k), \ \hat{Y}_{k1}, \hat{Y}_{k2}, \ldots, \hat{Y}_{kj} \in \hat{Y} \quad (2)$$

Our model, the sequence-spectrogram neural network $F$, accomplishes the classification task with the augmented data. The time-series samples are sequentially repeated to align with the corresponding synthetic images, which includes the original spectrograms and augmented spectrograms initially originated from the time-series. The merged samples $Q_{lj} = (X_l, \hat{Y}_{kj}) \in Q$ are imported into our model. The label predictions $\hat{L}_{lj}$ are procured afterwards:

$$\hat{L}_{lj} = F(Q_{lj}) \quad (3)$$

## III. METHODOLOGY

### A. Data Preparation and Augmentation

#### 1) Time-Frequency Transfer with STFT

Time-frequency image is taken into account with two primary purposes. Firstly, the spectral representation of the TSS contains abundant useful information from the frequency domain that couldn't be found in the whole series and vice versa. Therefore, we seek the time-series representation with both temporal and frequential features and eventually choose the STFT spectrogram as our augmentation method of the primitive TSS.

STFT as a time-frequency analysis technique, has been widely used in signal processing and spectral analysis [12]. It involves breaking down a time-domain signal into short, overlapping segments or windows and then computing the Fast Fourier Transform (FFT) for each window. By applying it to successive overlapping windows, a dynamic and time-varying representation of spectral characteristics is attained.

Mathematically, for a given input signal $x(t)$ at a timestamp $t$, the STFT is calculated as follows:

$$X(t, f) = \int_{-\infty}^{\infty} x(\tau) \cdot w(\tau - t) \cdot e^{-2\pi j f \tau} d\tau \quad (4)$$

where $X(t, f)$ denotes the spectral component or vector at each $t$ (window frame) and $f$, $w(n)$ represents the sliding window function and $f$ is the frequency variable. Typically, the STFT is more commonly applied to discrete signals, where it can be computed using the following formula:

$$X(m, f) = \sum_{n=0}^{N-1} x[n] \cdot w[n - mR] \cdot e^{-2\pi j f n} \quad (5)$$

where $N$ represents the signal's length, $R$ denotes the overlapping distance (numbers of points) between each adjacent sliding window and $m$ for the index of the window. After that, the spectrogram $S$ is generated through:

$$S = \log(|X(m, f)|) \quad (6)$$

We take the logarithm of the magnitude of $X(m, f)$ to yield the final spectrogram and set the $m = [m_1, m_2, \ldots, m_{max}]$ as the x label, the y label with the frequency magnitudes $f = [f_1, f_2, \ldots, f_{max}]$, each pixel with the value of the spectrogram is mapped from its corresponding $t$ and $f$. Then the values of the spectrogram are reset from 0 to 255 where a representation of grayscale image is produced.

#### 2) Image Augmentation through Random Erasing

To enlarge our datasets, random erasing is sifted through randomly selecting a rectangular region of an image and erasing its pixels with random values [13]. Given an image $I$ in our dataset with width $W$ and height $H$ along with a parameter of rectangular area $R_a$ ranging from [0.1,1], the random width $W_e$ of the rectangular occlusion is confined between $0.1W$ and $R_aW$, the height $H_e$ between $0.1H$ and $R_aH$. After that a random point $P = (x_e, y_e)$ is initialized. If $x_e + W_e < W$ and $y_e + H_e < H$, a region $I_e = [x_e, y_e, x_e + W_e, y_e + H_e]$ is set thereafter, otherwise the procedure above repeats until a felicitous $P$ is selected. Every pixel in the region $I_e$ is substituted with random value between [0, 255] to compose an erasing effect. The Fig. 1 displays an STFT image before and after random erasing.

Behind the selection for random erasing underlies two main reasons. Initially, as far as the texture features [14] of the spectrogram are concerned, which represent the information of the TF domain. Dissimilar to color transformation like brightness and contrast adjustment, which may cause severe changes of information within the spectrogram since the brightness of points reflects disparate degrees of STFT magnitude, random erasing merely appends an erasing rectangle, thereby reserving the remaining area with its original information. In addition, random erasing saves the location information of the remaining points, which correspond to their respective frequency and time coordinate positions. It fairly prevents the spectrogram from disordered and global permutation, which could take place through geometric transformations like cropping, scaling, rotation and flipping. Furthermore, the random erasing operation, for which it can be assumed as noise originating from real-world disturbance, is interpretable as a spontaneous process.

### B. Sequence-Spectrogram Neural Network (SSNN)

While employing the STFT spectrogram as a representation capturing both temporal and spectral characteristics, it is essential to emphasize that the original time series data should not be dismissed. The STFT spectrogram effectively encodes the frequency information through time order, yet it tends to inadequately capture intricate dynamic patterns within the sequence. Accordingly, the whole series is kept to pair with its corresponding spectrogram image as a multimodality fusion representation of a single TSS.

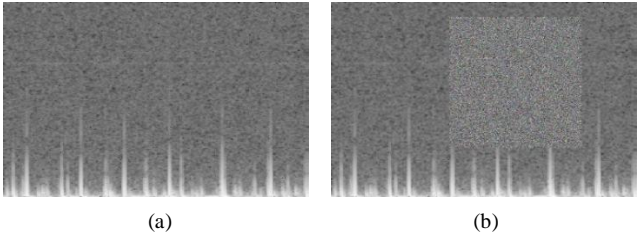

Figure 1. Augmentation process of STFT spectrogram. (a) Original STFT image, (b) Augmentation image after random erasing.

Our SSNN, shown in Fig. 2, is designed with two main courses, one as the features extractor of the time-series while the other one is to abstract the spatial features within the STFT spectrogram of time-frequency domain. The sequential module $F_1$, is mainly constructed with an initially convolutional layer assembled with a batch normalization layer, an ReLU as activation function and a maxpool layer. A string of 1-dimensional residual blocks follows by and each of which consists of two stacked convolutional layers and an identity mapping (shortcut) across them.

After the first convolutional layer, it lies a batch normalization layer and a ReLU. The next convolutional layer leaves the ReLU behind the addition operation set outside, which serves to add the input of the residual block to its output. Every block maintains the feature size of the input unchanged except the third one, who increases the amount of the channels and dwindles the size of the feature (downsampling). The shortcut part of the third block is constructed with a convolutional layer with a stride of 2, the same to the stride of its first convolutional layer in the main course, which is aimed at operating an aligned addition. The module ends up with a global average pool layer and a squeeze operation that transform the output features into a sequence.

Its counterpart, the frequency module $F_2$, is all the same construction but supplants the 1D convolutional layers with the 2D ones. The output is also transformed into a sequence with the same length as we weigh time-series and spectrogram the same. After extracting the temporal and the frequency features through the dual modules, both features are reduced into one dimension and concatenated into a whole sequence of features along the channel axis, where information from two modalities is conflated. Finally, linear layers are added at the end of our model assembling the fully connected layer, which provides the possibility distribution as a prediction $D$. Within our fusion model, $F_1$ is fed with time-series $X$ whereas $F_2$ receives the augmented spectrograms, the greyscale images $Y$:

$$D = \text{Linear}\{\text{Concat}[F_1(X), F_2(Y)]\} \quad (7)$$

The final predicted label is determined through a function $argmax$ that choose the class $C_i \in C$ which has the largest possibility in $D$:

$$C_i = argmax(D) \quad (8)$$

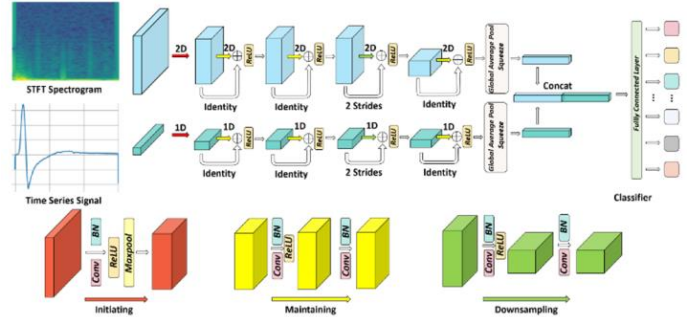

Figure 2. The overall structure of the proposed SSNN model.

To train our proposed SSNN model, the loss function is utilized by considering the categories to be classified. The BCEWithLogits loss is calculated through the following formula with the $y_{C_i}$ as the true label (0 or 1) for the class $C_i$ and the $x_{C_i}$ as the output logits associated with the same class $C_i$ from the model:

$$loss(x, y) = -[\sum_{C_i}^{N} y_{C_i} \cdot \log(Sigmoid(x_{C_i})) + (1 - y_{C_i}) \cdot \log(1 - Sigmoid(x_{C_i}))] \quad (9)$$

where $N$ represents the number of classes and the $Sigmoid$ is the activation function to squash the logits $x_{C_i}$ into the range $[0,1]$, which is interpreted as the estimated probability of the sample belonging to the class $C_i$.

## IV. EXPERIMENT RESULTS

### A. Dataset Setup

#### 1) The Amyotrophic Lateral Sclerosis Dataset

In our investigation, we source the Amyotrophic Lateral Sclerosis (ALS) dataset from [15] arbitrarily, which comprises a total of 300 samples recorded from the biceps and medial vastus muscle and then is bisected into two classes: ALS (150) and Normal (150) derived from 10 normal subjects (4 females and 6 males) aged 21-37 years and 8 patients (4 females and 4 males) aged 35-67 years. Every EMG signal represents a 1-second record fragmented from an approximate 11-second record sampled at 24kHz. We split our ALS dataset into a training set with an amount of 80% (240) and a test set with 20% (300) at random.

For the image dataset, the dataset above is transferred into the STFT spectrogram image by setting the window length $l$ as 420 and overlapping size $R$ as $0.75l$. For the augmented dataset, only the trainset is augmented through random erasing by setting the rectangular area $R_a$ as $[0.5, 0.6, 0.7]$ for each sample; therefore, with the original samples, the number of total augmented samples of the trainset reaches 960. For the fusion dataset, time-series samples are aligned to the corresponding STFT image samples through repetitions over three times, resulting in each sample incorporating two distinct modalities. The consecutive transformations of the ALS dataset are shown in the Fig. 3.

#### 2) The Wind Turbine Fault Dataset

The WTF dataset consisting of 4 classes: Crack, Broken, Normal and Faring. Housing 200 samples in total and each class with 50, this dataset is split sequentially over the ratio of trainset and the test set as 2:1. Each WTF signal comprises 9 channels from three sensors (each of which has 3 axis) located on the wind turbine.

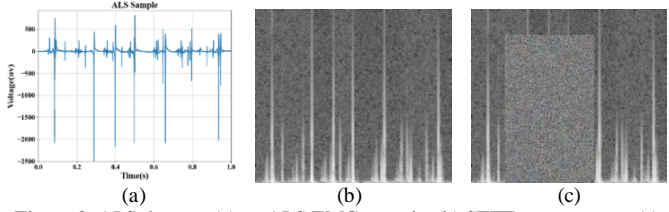

Figure 3. ALS dataset. (a) an ALS EMG sample, (b) STFT spectrogram, (c) augmentation with random erasing.

Similar to 1), we choose the window length $l$ as 62 and overlapping size $R$ as $0.91l$ to transform the multivariant time-series into TF domain images with the same quantity and select the rectangular area $R_a$ of random erasing with $[0.3, 0.4, 0.5, 0.6, 0.7]$ to generate augmented STFT pictures. Thus, with each sample containing 9 STFT images, a trainset of 792 samples in total is obtained. We deliberately construct this WTF dataset by pairing each temporal variant with the corresponding TF variant (augmented) in one sample. The Fig. 4 has shown a transitional process of a single channel (time-series) from one of the sensors on the wind turbine to an augmented STFT picture.

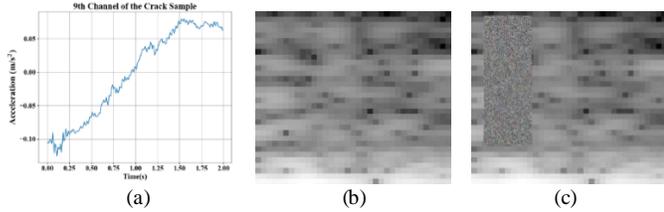

Figure 4. WTF dataset. (a) an acceleration signal of the third sensor on the z axis, (b) STFT spectrogram, (c) augmentation with random erasing.

## B. Comparison Experimental Results

### 1) Experiments set

To validate the performance of the proposed model, extensive experiments are conducted and compared with other state-of-the-art models in three conditions as follows: (1) Time-series representation: Our first protocol employs 3 well-behaved models including ResNet, GRU and FCN to provide a baseline performance of solely using the temporal modality, the whole series. For the ALS dataset, we specifically use the ALSNet from [15] to provide a better oriented benchmark. (2) Spectrogram representation: ResNet18 [16], a model designed exclusively for image classification is taken into account to examine the efficacy of TF transformation. In addition, augmented image data is also trained with the same test set to verify whether the random erasing augmentation contributes to the few-shot learning problem. (3) Fusion representation: A fusion model MedFuse [17], as a benchmark, along with our model are trained with the fusion dataset from both time and TF domains with or without data augmentation.

### 2) Evaluation Metrics

Metrics included are as follows: F1 score is used to evaluate the performance of a binary classification model, combining both precision and recall into a single value and providing a harmonic measure of the model's accuracy:

$$F1 = 2\frac{precision \cdot recall}{precision + recall} = \frac{TP}{TP + \frac{1}{2}(FP + FN)} \quad (10)$$

Macro F1-score is an extension of the F1-score for multi-class classification problems. It calculates the F1-score independently for each class and then takes the unweighted average of the individual F1-scores:

$$\text{MF1} = \frac{1}{C}\sum_{i=1}^{C} F1 \quad (11)$$

Accuracy measures the proportion of correctly classified instances over the total instances, directly representing the performance of models on the TSC:

$$\text{Accuracy} = \frac{TP + TN}{TP + TF + FP + FN} \quad (12)$$

### 3) Experimental Results

As seen in Table 1, the proposed SSNN model achieves the best scores based on the augmented ALS dataset with 93.75% F1 score and 93.33% accuracy, showcasing our advance in TSC few-shot problem. Within the ALS dataset without augmentation, the proposed model still achieves relatively the best scores with 87.27% F1 score and 88.33% accuracy. With the data augmentation, the accuracy and F1 score of ResNet18 are elevated by 3.33% and 3.07%, MedFuse by 3.33% and 3.39%, and SSNN by 5% and 6.48% with the most improvement. Compared to the unaugmented ALS dataset, all results of the above three models have been ameliorated, indicating that our measure of random easing to enlarge the ALS dataset has contributed to the performance of models. Additionally, it is notable that ResNet18 overshadows MedFuse in both cases but with slight advantages. Compared to the aforementioned models, simple time-series models could hardly transcend the others, and the best one is the ALSNet, which is equal to the MedFuse on accuracy (85%). Among all the models, GRU reaches the lowest score with the accuracy of merely 70%.

As shown in Table 2, our model still attains the best scores based on the augmented WTF dataset with the MF1 score of 95.48% and the accuracy of 95.59%. The second-best model is the FCN based on the unaugmented WTF dataset with the modality of time series, which achieves the MF1 score of 89.23% and the accuracy of 89.71%. It is noteworthy that the FCN surpasses our proposed model using the unaugmented WTF dataset (fusion) with the MF1 score by 2.33% and the accuracy by 2.95%. Similarly, the performance of ResNet18 and MedFuse has been improved when we implement the augmented WTF dataset, with the MF1 score increasing 6.99% and 7.37%, with accuracy raising 5.88% and 7.35%. However, ResNet18 (STFT image) is relatively worse than MedFuse (Fusion) with respect to the augmented or the unaugmented dataset. Still, the evaluation metrics manifest that GRU obtains the worst performance with the MF1 score of 73.54% and the accuracy of 75%.

Fig. 5 further shows the classification performance by confusion matrix using the proposed SSNN model. Without data augmentation, the confusion matrix is shown in Fig. 5 (a) with the ALS detection rate of the same 96.7% and the true negative rate of 80%. With data augmentation, the confusion matrix is shown in Fig. 5 (b) with the ALS detection rate of 96.7% and the true negative rate of 90%. The confusion matrix results of unaugmented WTF dataset is represented in Fig. 5 (c). The detection rates of Crack and Faring are both 100% while Broken and Normal are lower than SSNN based on the augmented WTF dataset by 11.8% and 23.5%, respectively. The confusion matrix of augmented WTF dataset is presented in Fig. 5 (d) with detection rates of Crack, Faring, Normal as 100% and of Broken as 82.4%.

TABLE I. COMPARISON STUDY OF ALS DATASET

| Compared Model | Without Augmentation | | With Augmentation | |
|---|---|---|---|---|
| | F1(%) | Accuracy(%) | F1(%) | Accuracy(%) |
| ResNet[4] | 79.98 | 80.00 | NA | NA |
| GRU[5] | 70.00 | 70.00 | NA | NA |
| FCNs[6] | 76.36 | 78.33 | NA | NA |
| ALSNet[15] | 84.21 | 85.00 | NA | NA |
| ResNet18[16] | 86.21 | 86.67 | 89.28 | 90.00 |
| MedFuse[17] | 84.75 | 85.00 | 88.14 | 88.33 |
| Ours | **87.27** | **88.33** | **93.75** | **93.33** |

NA: Not available.

TABLE II. COMPARISON STUDY OF WTF DATASET

| Compared Model | Without Augmentation | | With Augmentation | |
|---|---|---|---|---|
| | MF1(%) | Accuracy(%) | MF1(%) | Accuracy(%) |
| ResNet[4] | 85.43 | 85.29 | NA | NA% |
| GRU[5] | 73.54 | 75.00 | NA | NA |
| FCNs[6] | 89.23 | 89.71 | NA | NA |
| ResNet18[16] | 72.21 | 73.53 | 79.20 | 79.41 |
| MedFuse[17] | 78.06 | 77.94 | 85.43 | 85.29 |
| Ours | 86.90 | 86.76 | **95.48** | **95.59** |

NA: Not available.

The entire experimental results show that our proposed model SSNN is accurate and feasible for the TSC few-shot learning problem. Moreover, it shows that the augmentation using random erasing can improve the classification performance while the feature fusion of the temporal and TF domain is robust to both univariate and multivariate datasets.

## V. CONCLUSIONS

In this paper, a novel deep learning method for TSC few-shot learning was proposed. The proposed SSNN model integrated a time-frequency transformation and data augmentation processes while combined with the original time series inputs. Extensive experiments were conducted to show the superiority of our approach, which provided the highest accuracy and F1 score in both datasets. In the future work, we plan to focus on more representative TF transformation approaches and more robust augmentation methods that can felicitously deal with the TSC few-shot learning problems.

ACKNOWLEDGEMENT

This work is funded by Shandong Science Fund for Excellent Overseas Young Scholars (2023HWYQ-005).

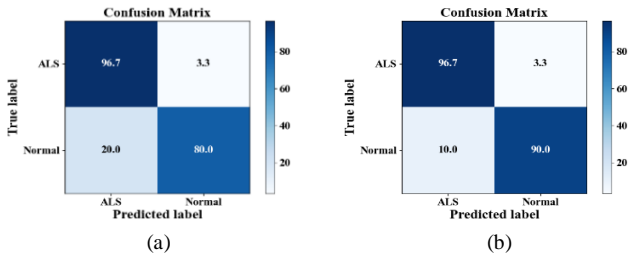

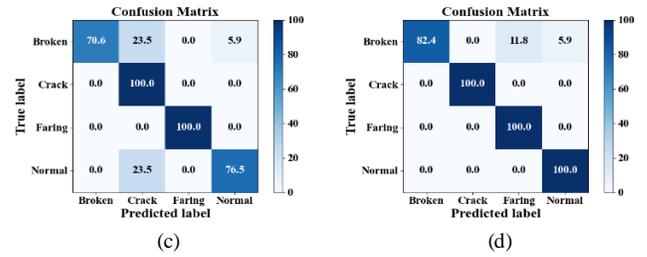

(c) (d)

*Figure 5: Confusion matrix. (a) unaugmented ALS dataset (b) augmented ALS dataset (c) unaugmented WTF dataset (d) augmented WTF dataset.*